\documentclass[letterpaper, 10 pt, conference]{ieeeconf}  

\IEEEoverridecommandlockouts                              

\usepackage{amsmath,amssymb,amsfonts}
\usepackage[ruled,vlined]{algorithm2e}
\usepackage{textcomp}
\usepackage{xcolor}
\usepackage{array}
\usepackage[normalem]{ulem}
\usepackage{tabularray}
\usepackage{multirow}
\usepackage{threeparttable}
\usepackage[caption=false,font=normalsize,labelfont=sf,textfont=sf]{subfig}
\usepackage{stfloats}
\usepackage{siunitx}
\usepackage{stmaryrd}
\usepackage{url}
\usepackage{verbatim}
\usepackage{graphicx}
\usepackage{cite}
\usepackage{caption}
\usepackage{color}
\usepackage{colortbl}
\usepackage{booktabs}
\usepackage{hyperref}
\let\OLDthebibliography\thebibliography
\renewcommand\thebibliography[1]{
  \OLDthebibliography{#1}
  \setlength{\parskip}{0pt}
  \setlength{\itemsep}{0pt plus 0.3ex}
}
\title{\LARGE \bf
VVLoc: Prior-free 3-DoF Vehicle Visual Localization
}

\author{Ze Huang$^{1}$, Zhongyang Xiao$^{2, \dagger}$, Mingliang Song$^{2}$, Longan Yang$^{2}$, Hongyuan Yuan$^{2}$, and Li Sun$^{3}$
\thanks{This work was fulfilled when Ze Huang interned at NIO.}
\thanks{$^{1}$ Fudan University. 
$^{2}$ NIO. 
$^{3}$ Bosch XC.}
\thanks{$^{\dagger}$ Corresponding author \texttt{xiaozhongyang@yeah.net}}
}

\begin{document}

\maketitle
\thispagestyle{empty}
\pagestyle{empty}

\begin{abstract}

Localization is a critical technology in autonomous driving, encompassing both topological localization, which identifies the most similar map keyframe to the current observation, and metric localization, which provides precise spatial coordinates.
Conventional methods typically address these tasks independently, rely on single-camera setups, and often require additional 3D semantic or pose priors, while lacking mechanisms to quantify the confidence of localization results, making them less feasible for real industrial applications.
In this paper, we propose VVLoc, a unified pipeline that employs a single neural network to concurrently achieve topological and metric vehicle localization using multi-camera system. 
VVLoc first evaluates the geo-proximity between visual observations, then estimates their relative metric poses using a matching strategy, while also providing a confidence measure.
Additionally, the training process for VVLoc is highly efficient, requiring only pairs of visual data and corresponding ground-truth poses, eliminating the need for complex supplementary data.
We evaluate VVLoc not only on the publicly available datasets, but also on a more challenging self-collected dataset, demonstrating its ability to deliver state-of-the-art localization accuracy across a wide range of localization tasks.

\end{abstract}

\section{Introduction}
Localization is a fundamental technology in autonomous driving, underpinning critical tasks such as global map reconstruction, map-fusion-based perception, and sensor calibration. 
The increasing adoption of the Bird's-Eye View (BEV) representations has spurred the development of novel approaches to localization, particularly using multi-camera systems.
Recent studies have begun to explore the application of visual BEV in localization tasks, as summarized in Tab.\ref{tab:tax}.
Some of these approaches\cite{camiletto2023u, shore2023bev, ge2024bev2pr} focus on topological localization, leveraging BEV representations to enhance overall precision by incorporating tasks like map segmentation during neural network training.
Other methods\cite{zhang2022bev, he2023egovm, zhao2022jperceiver, li2023occ, wu2024maplocnet} have explored visual BEV for metric localization or attempted to unify topological and metric localization\cite{fervers2023c, xu2023leveraging}. 
However, due to scale ambiguity in camera systems, several methods\cite{zhang2022bev, he2023egovm, fervers2023c, wu2024maplocnet} are confined to geometric maps with predefined scales, such as vectorized maps, which limits their versatility. 
In contrast, visual-to-visual metric localization approaches\cite{zhao2022jperceiver, li2023occ} often require explicit 3D supervision to mitigate scale uncertainty, and without it, accurate metric translation remains difficult~\cite{xu2023leveraging}. 
Moreover, these methods generally lack confidence estimation for localization results, further hindering their reliability in practical deployment.

\begin{table}
\centering
\resizebox{0.48\textwidth}{!}{
\begin{tabular}{l|c|c|ccc} 
\toprule
\textbf{Approach} & \textbf{Map} & \textbf{Other Supervision} & \textbf{TL} & \textbf{ML} & \textbf{Conf.} \\ 
\hline
BEV-locator~\cite{zhang2022bev} & {\cellcolor[HTML]{FCE4D6}Vect.} & {\cellcolor[HTML]{E2EFDA}-} & {\cellcolor[HTML]{FCE4D6}$\times$} & {\cellcolor[HTML]{E2EFDA}3-DoF} & {\cellcolor[HTML]{FCE4D6}$\times$} \\
EgoVM~\cite{he2023egovm} & {\cellcolor[HTML]{FCE4D6}Vect.} & {\cellcolor[HTML]{FCE4D6}M} & {\cellcolor[HTML]{FCE4D6}$\times$} & {\cellcolor[HTML]{E2EFDA}3-DoF} & {\cellcolor[HTML]{E2EFDA}$\surd$} \\
U-BEV~\cite{camiletto2023u} & {\cellcolor[HTML]{FCE4D6}SD} & {\cellcolor[HTML]{FCE4D6}M} & {\cellcolor[HTML]{E2EFDA}$\surd$} & {\cellcolor[HTML]{FCE4D6}-} & {\cellcolor[HTML]{FCE4D6}-} \\
MapLocNet~\cite{wu2024maplocnet} & {\cellcolor[HTML]{FCE4D6}Rast.} & {\cellcolor[HTML]{FCE4D6}M} & {\cellcolor[HTML]{FCE4D6}$\times$} & {\cellcolor[HTML]{E2EFDA}3-DoF} & {\cellcolor[HTML]{FCE4D6}$\times$} \\
BEV-CV~\cite{shore2023bev} & {\cellcolor[HTML]{FCE4D6}Aerial} & {\cellcolor[HTML]{FCE4D6}M} & {\cellcolor[HTML]{E2EFDA}$\surd$} & {\cellcolor[HTML]{FCE4D6}-} & {\cellcolor[HTML]{FCE4D6}-} \\
C-BEV~\cite{fervers2023c} & {\cellcolor[HTML]{FCE4D6}Aerial} & {\cellcolor[HTML]{E2EFDA}-} & {\cellcolor[HTML]{E2EFDA}$\surd$} & {\cellcolor[HTML]{E2EFDA}3-DoF} & {\cellcolor[HTML]{E2EFDA}$\surd$} \\ 
\hline
JPerceiver~\cite{zhao2022jperceiver} & {\cellcolor[HTML]{E2EFDA}Visual} & {\cellcolor[HTML]{FCE4D6}M, D} & {\cellcolor[HTML]{FCE4D6}$\times$} & {\cellcolor[HTML]{E2EFDA}6-DoF} & {\cellcolor[HTML]{FCE4D6}$\times$} \\
OCC-VO~\cite{li2023occ} & {\cellcolor[HTML]{E2EFDA}Visual} & {\cellcolor[HTML]{FCE4D6}O} & {\cellcolor[HTML]{FCE4D6}$\times$} & {\cellcolor[HTML]{E2EFDA}6-DoF} & {\cellcolor[HTML]{FCE4D6}$\times$} \\
BEV2PR~\cite{ge2024bev2pr} & {\cellcolor[HTML]{E2EFDA}Visual} & {\cellcolor[HTML]{FCE4D6}M} & {\cellcolor[HTML]{E2EFDA}$\surd$} & {\cellcolor[HTML]{FCE4D6}-} & {\cellcolor[HTML]{FCE4D6}-} \\
vDISCO~\cite{xu2023leveraging} & {\cellcolor[HTML]{E2EFDA}Visual} & {\cellcolor[HTML]{E2EFDA}-} & {\cellcolor[HTML]{E2EFDA}$\surd$} & {\cellcolor[HTML]{E2EFDA}1-DoF} & {\cellcolor[HTML]{FCE4D6}$\times$} \\ 
\hline
\textbf{VVLoc (Ours)} & \cellcolor[HTML]{E2EFDA}Visual & \cellcolor[HTML]{E2EFDA}- & \cellcolor[HTML]{E2EFDA}$\surd$ & \cellcolor[HTML]{E2EFDA}3-DoF & \cellcolor[HTML]{E2EFDA}$\surd$ \\
\bottomrule
\end{tabular}}
\caption{\textbf{Task Comparison and Taxonomy.} `Vect.' denotes vectorized map, `SD' semantic-dense map, `Rast.' rasterized map, `Aerial' aerial map, and `Visual' visual keyframe map. \textbf{M} stands for map segmentation, \textbf{D} depth estimation, and \textbf{O} occupancy prediction. \textbf{TL} and \textbf{ML} indicate topological and metric localization capability. \textbf{Conf.} denotes confidence scores used to filter unreliable metric localization results.}
\label{tab:tax}
\vspace{-4em}
\end{table}

In summary, existing methods rarely achieve unified topological and metric localization simultaneously. 
Those that attempt to do so typically rely on additional supervision, which is labor-intensive, or require high-quality scaled map, limiting their real-world applicability. 
These constraints underscore the need for a more flexible and resource-efficient solution to vehicle visual localization.
To address these challenges, we propose VVLoc, a unified visual localization framework that performs both topological and metric localization.
To tackle the challenge of scale ambiguity inherent in camera-based system, we introduce novel components to extract local-view descriptors from multi-camera inputs and leverage relative pose supervision for descriptor learning. 
Combined with an iterative pose estimation module based on descriptor matching, our method achieves accurate and scaled metric pose estimation without explicit scale constraints, while simultaneously providing confidence estimates for the localization results.
VVLoc requires only ground-truth pose supervision during training, offering a more practical and scalable solution for real-world deployment.

The main contributions are:
\textbf{(i)} We propose a novel vehicle visual localization method that unifies topological and metric localization, eliminating the need for prior poses, multi-frame observation, and additional supervision, beyond ground-truth poses during training.
\textbf{(ii)} We develop a matching-based pose estimation method tailored for visual input, which alleviates the scale ambiguity challenge in driving scenarios while enhancing topological localization accuracy.
\textbf{(iii)} Beyond public datasets, we further collected a challenging real-world dataset and designed a comprehensive evaluation protocol covering global localization, loop closure detection, pose estimation, and multi-model mapping to validate our method.

\begin{figure*}[htb]
  \centering
  \includegraphics[width=1.0\linewidth]{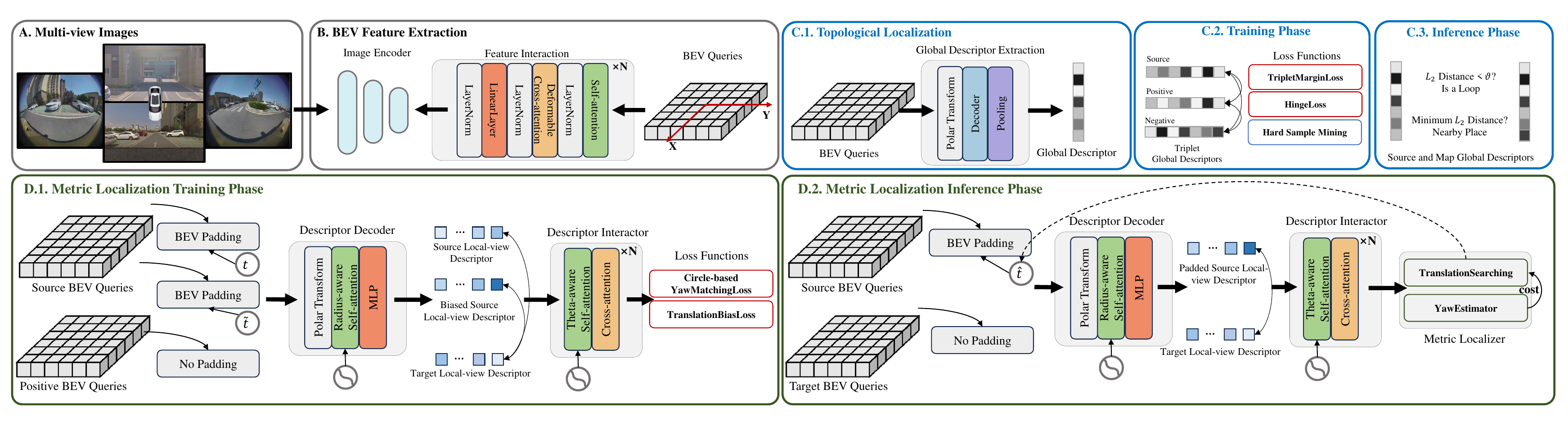}
   \caption{\textbf{Pipeline of the VVLoc:} For the input \textbf{(A)} multi-camera images, we leveraging \textbf{(B)} BEVformer\cite{li2022bevformer} for BEV queries extraction and \textbf{(C)} decode global descriptor from the BEV queries for topological localization. \textbf{(D)} For metric localization, we decode BEV queries into local-view descriptors and combine BEV Padding for metric pose estimation.}
   \label{fig:VVLoc}
\vspace{-1.5em}
\end{figure*}

\section{Related Works}
\noindent{\bf Visual Topological Localization} 
identifies the map frame corresponding to the source frame. i.e., current observation. 
The early methods\cite{arandjelovic2016netvlad, jin2017learned, radenovic2018fine} used CNN backbones and global descriptors generated via pooling layers for frame retrieval. 
Patch-NetVLAD\cite{hausler2021patch} introduced patch-level local feature matching for re-ranking candidates. 
Recent advancements\cite{lu2024cricavpr, tzachor2024effovpr} utilize foundation models pretrained on extensive datasets, to enhance accuracy. 
Despite progress, research predominantly targets monocular setups, lacks solutions for multi-camera systems, and emphasizes retrieval metrics over practical integration with downstream tasks.

\noindent{\bf Visual Pose Estimation} further determines the relative pose between source and map frames. 
The early approaches\cite{detone2018superpoint, barroso2019key} followed traditional keypoint detection and descriptor extraction.
Later methods improved robustness via joint keypoint extraction\cite{revaud2019r2d2, gleize2023silk}, detector-free approaches\cite{sun2021loftr, jiang2021cotr}.
Dense methods\cite{edstedt2023dkm, edstedt2024roma}  estimate a dense warp by regressing pixel coordinates, aiming to match every possible pixel pair between images. 
Few studies address metric pose estimation directly, with regression-based methods\cite{winkelbauer2021learning, cai2021extreme} often unreliable due to the absence of confidence measures. 
Recently, MicKey\cite{barroso2024matching} offers a promising alternative, estimating scaled poses via self-supervised feature matching.

\noindent{\bf Visual BEV for Localization}
We have cataloged all existing works that involve visual BEV for localization in Tab.~\ref{tab:tax}. 
Recent works\cite{camiletto2023u, shore2023bev, ge2024bev2pr} leverage BEV representations and 3D semantic supervision to improve topological localization, albeit with high training costs and marginal benefits\cite{ge2024bev2pr}. 
Metric localization efforts\cite{zhang2022bev, he2023egovm, fervers2023c, wu2024maplocnet} rely on scale-clear maps, which limit data collection and require accurate initial poses.
Visual-to-visual metric localization often depends on complex 3D supervision\cite{li2023occ, zhao2022jperceiver}, while methods like vDISCO\cite{xu2023leveraging} forego 3D data but are restricted to relative yaw estimation, making them insufficient for broader localization tasks.
Moreover, most visual metric localization methods lack a confidence score to assess pose reliability, raising concerns for practical deployment.

\section{Methodology}
\noindent{\bf Problem Definition:} 
Given a single frame of multi-camera images $I_{s} = \left \{ I^{i}_{s}\mid i=1, 2, ...,M \right \}$ refer to as source images, the goal of VVLoc is: \textbf{(i)} Determine whether map keyframe $I_{t}$ is close to $I_{s}$' s geographical location; \textbf{(ii)} If so, estimate the metric relative pose $\hat{\xi}_{s \rightarrow t} = (\hat{x}_{s\rightarrow t}, \hat{y}_{s\rightarrow t}, \hat{\varphi}_{s\rightarrow t})$ between the two frames. 
Only ground-truth poses and calibrated camera parameters are provided during training.

\noindent{\bf Visual BEV Encoder} The BEV encoder serves to extract features of the multi-camera images and project them to BEV space.
We treated this part in the same way as the BEVformer\cite{li2022bevformer}, shown in Fig.\ref{fig:VVLoc}.B.
The extracted versatile BEV queries $Q \in \mathbb{R}^{H \times W \times C}$ is responsible for the corresponding grid cell region in the BEV plane.
Each grid cell corresponds to a real-world size of $g$ meters in ego-vehicle system.
BEV queries will be further used to decode task-specific descriptors.

\subsection{Visual Topological Localizer}

\noindent{\bf Global Descriptor Decoder}
The topological localization is based on global descriptor comparison, shown in Fig.\ref{fig:VVLoc}.C.
Initially, we use polar transformation to transform the \textit{Cartesian} BEV queries $Q \in \mathbb{R}^{H \times W \times C}$ into polar BEV queries $\mathcal{Q} \in \mathbb{R}^{T \times R \times C}$, where $T$ is the angular samples and $R$ is the radial samples. 
This operation aims to align the feature format with that used in the subsequent metric localization, and also reduce the feature spatial dimensionality.
Finally, VVLoc leveraging feature pooling approach to aggregate the polar BEV queries into global descriptors.

\noindent{\bf Topological Localizer}
Topological localization is achieved by comparing the \textit{L2}-distances between the normalized global descriptors of the source image $\mathcal{G}_{s}$ and the map images  $\left \{\mathcal{G}_{k} \mid k=1,2,..,N \right \}$.
For global localization, the $K$ closest map descriptors to $\mathcal{G}_{s}$  are selected as candidates  $\mathcal{C} = \left \{\mathcal{G}_{k} \mid k \in \text{argsort}\left( \|\mathcal{G}_{s} - \mathcal{G}_{k}\|_2 \right)_{1:K} \right \}$, which are then refined using a re-ranking strategy.
For loop closure detection, candidates are map descriptors within a distance threshold $\vartheta_{p}^{h}$, $\mathcal{C} = \left \{\mathcal{G}_{k} \mid \|\mathcal{G}_{s} - \mathcal{G}_{k}\|_2 < \vartheta_{p}^{h} \right \}$.
Metric relative poses for these candidates are estimated using the metric localizer.

\begin{figure}[htb]
  \centering
  \includegraphics[width=1.0\linewidth]{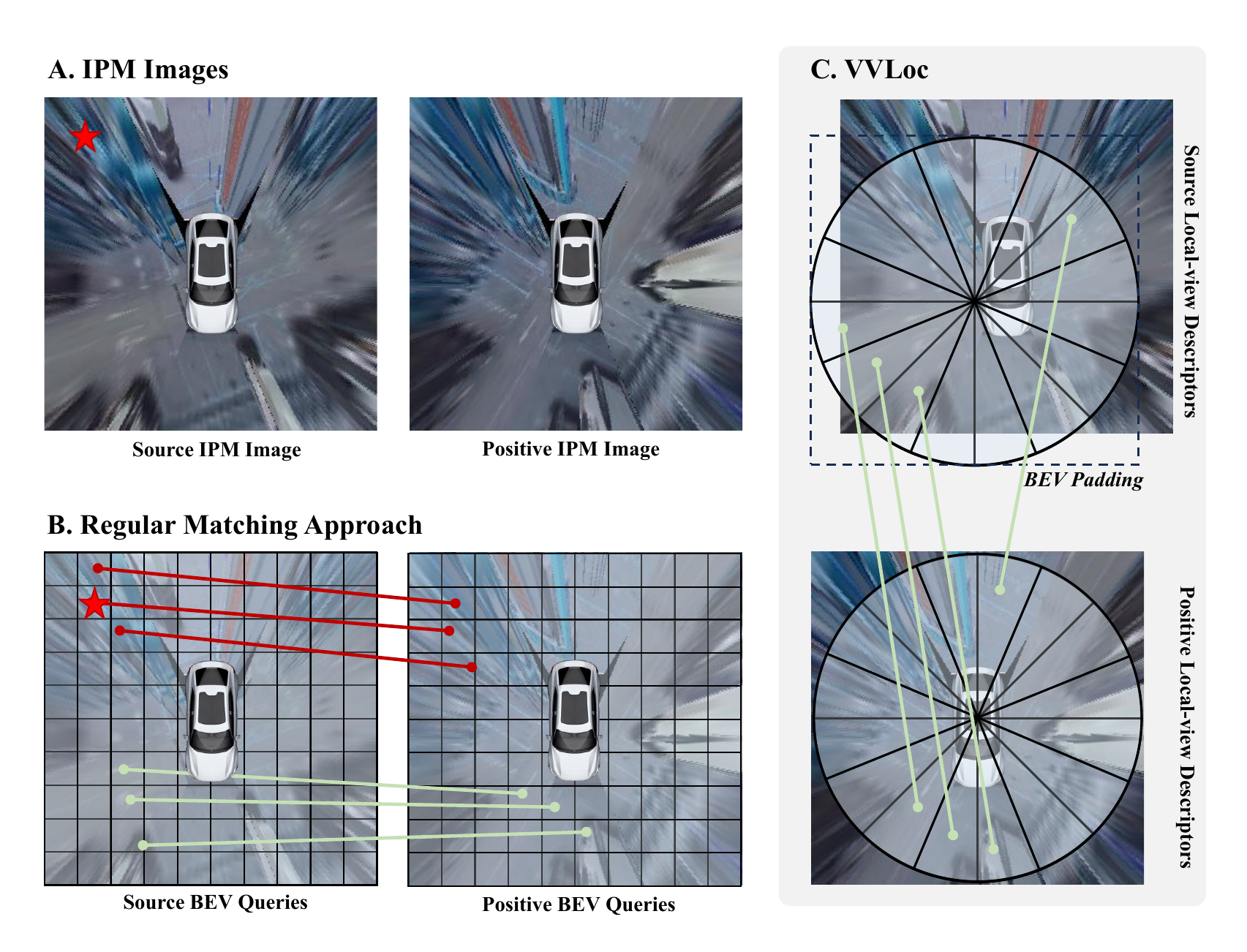}
   \caption{\textbf{Matching Strategy Comparison.} Conventional matching-based strategies produce ambiguous supervision in visual BEV, which hinders model training. VVLoc addresses this issue by introducing a method that leverages BEV Padding and local-view descriptor matching to estimate the metric relative pose accurately.}
   \label{fig:idea}
\vspace{-2em}
\end{figure}

\subsection{Visual Metric Localizer}
\label{sec: ml}
\noindent{\bf Core Idea} A direct metric localization approach that also provides confidence estimation can be inspired by point cloud registration~\cite{huang2021predator, qin2023geotransformer}, where BEV queries are matched and the relative pose is solved in ego-vehicle coordinates.
However, this can lead to ambiguity, as shown in Fig.\ref{fig:idea}.A, where the visual BEV queries does not reflect the actual geometric position but instead corresponds to image features along the line from that point to a camera. 
When applying point cloud registration, ambiguous supervision (Fig.\ref{fig:idea}.B) can complicate pose estimation. 
To address this, we propose a novel approach tailored for visual BEV. 
The key idea is to extract local-view descriptors from the two BEV queries, match them to estimate rotation, and reduce translation's impact by enumerating possible translations then applying BEV Padding. 
The final pose is determined by selecting the translation and rotation with the lowest matching cost.

\begin{figure}[htb]
  \centering
  \includegraphics[width=1.0\linewidth]{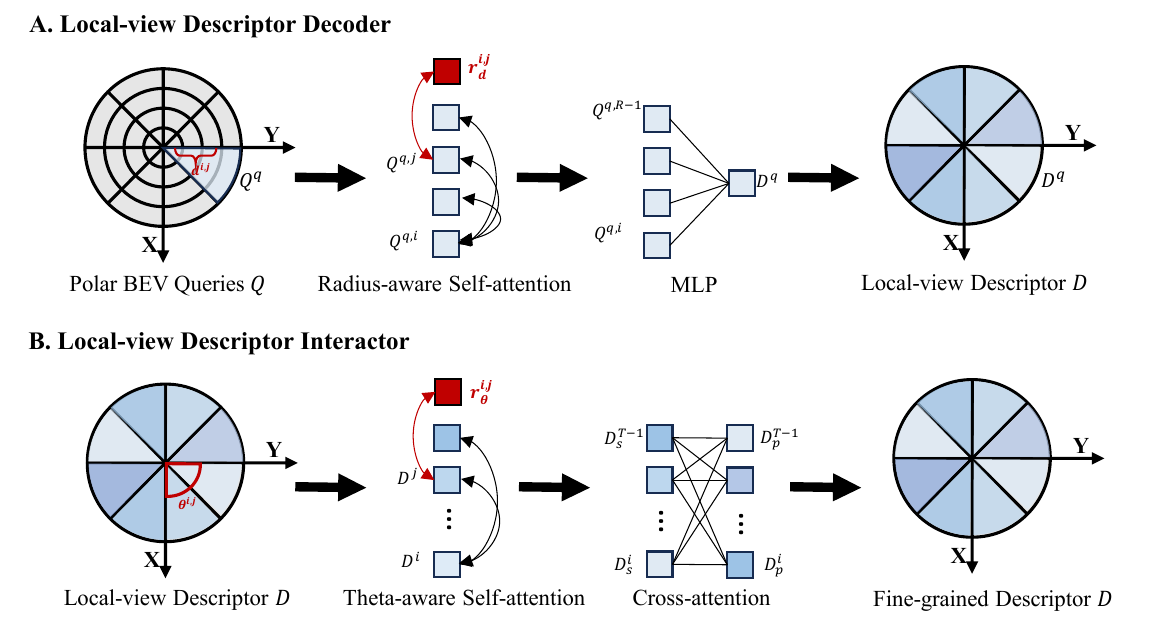}
   \caption{\textbf{The Proposed Novel Components.} \textbf{(A)} Radius-aware Self-attention and \textbf{(B)} Theta-aware Self-attention are implemented to enhance the geometric awareness of the local-view descriptors.}
   \label{fig:att}
\vspace{-1em}
\end{figure}

\noindent{\bf Local-view Descriptor Decoder}
The process of local-view descriptor extraction (Fig.\ref{fig:VVLoc}.D) begins with the polar BEV queries $\mathcal{Q} \in \mathbb{R}^{T \times R \times C}$. 
We will change this polar BEV queries into local-view descriptors $\mathcal{D} \in \mathbb{R}^{T \times C}$, capturing features from various view angles. 
As shown in Fig.\ref{fig:idea}.C, each descriptor in $\mathcal{D}$ geometrically represents the features within a field-of-view angle of $\frac{360}{T}$° from the positive y-axis to the positive x-axis, with the ego-vehicle at the center as the observer. 
To accomplish this change, we integrate the $R$ C-dimensional features that represent the radius dimension of the polar coordinates into a single C-dimensional feature. 
To effectively capture the geometric relationships in the polar BEV queries, we introduce a novel module, Radius-aware Self-attention ($\mathrm{RASA}$, details in Fig.\ref{fig:att}.A), which combines multi-head self-attention with geometric relative positional embedding. 
Specifically, we additionally add a geometric structure embedding when calculating attention score $e^{q, i, j}$ between $\mathcal{Q}^{q, i}$ and $\mathcal{Q}^{q, j}$:
\begin{equation}
    e^{q, i, j} = \frac{1}{\sqrt{C}}(\mathcal{Q}^{q, i}\mathrm{W}^{Q})(\mathcal{Q}^{q, j}\mathrm{W}^{K} + r_{d}^{i,j}\mathrm{W}^{R})^T.
\end{equation}
Here, $\mathrm{W}^{Q}$, $\mathrm{W}^{K}$, $\mathrm{W}^{R} $ are
the respective projection matrices.
$r_{d}^{i, j} \in \mathbb{R}^{C}$ is the distance-percepted geometric embedding, calculated using sinusoidal embedding function:
\begin{equation}
r^{i, j}_{2k}  = \sin (\frac{d_{i, j}}{\sigma_{r} 10000^{\frac{2k}{C} }} ), \quad r^{i, j}_{2k+1}  = \cos (\frac{d_{i, j}}{\sigma_{r} 10000^{\frac{2k}{C} }}),
 \label{eq:emb}
\end{equation}
where $\sigma_{r}$ is a temperature factor, $d_{i, j}$ is the coordinate $Euclidian$ distance corresponding to  $\mathcal{Q}^{q, i}$ and $\mathcal{Q}^{q, j}$ in ego-vehicle system.
Finally, we get local-view descriptor $\mathcal{D}  \in \mathbb{R}^{T \times C}$ by a multi-layer perceptron, and the whole process can be formulated as $\mathcal{D} = \mathrm {MLP}(\mathrm{RASA}(\mathcal{Q}))$.

\noindent{\bf Local-view Descriptor Interactor}
We further perform feature interacting on $ \mathcal{D}_{s}$ and  $ \mathcal{D}_{p}$ before matching, $\mathcal{D}_{s}, \mathcal{D}_{p} = \mathrm{MHCA}(\mathrm{TASA}( \mathcal{D}_{s}),  \mathrm{TASA}( \mathcal{D}_{p}))$.

Here, $\mathrm{MHCA}$ is the multi-head cross-attention, and $\mathrm{TASA}$ is the Theta-aware Self-attention (Fig.\ref{fig:att}.B) implemented using the same strategy as the proposed Radius-aware Self-attention.
Simply put, we add an angle-percepted geometric embedding $r_{\theta}^{i, j}$ when calculating attention score $e^{i, j}$ between $\mathcal{D}^{i}$ and $\mathcal{D}^{j}$.
The embedding calculation is same with Eq.\ref{eq:emb} but using another temperature factor $\sigma_{a}$.
$d_{i, j}$ is replaced by $\theta_{i, j}$, which refers to the relative angle corresponding to  $\mathcal{D}^{i}$ and $\mathcal{D}^{j}$ in ego-vehicle system.
After several rounds of feature interaction, the fine-grained local-view descriptor $\mathcal{D}_{s}, \mathcal{D}_{p}$ will be normalized and ready for matching.

\noindent{\bf Yaw Estimator}
With the fine-grained local-view descriptors represent each view angle range, we can simply estimate the relative rotation (yaw angle) $ \hat{\varphi}$ between $\mathcal{D}_{s}$ and $\mathcal{D}_{p}$ by minimizing the descriptor matching cost:
\begin{equation}
\begin{split}
\hat{\varphi} = \mathrm{REst}(\mathcal{D}_{s}, \mathcal{D}_{p})= \frac{-360}{T} \min_{ \varphi} (\delta(\rho(\mathcal{D}_{s}, \varphi, 0) , \mathcal{D}_{p})).
\end{split}
\end{equation}
Here, $\delta(A, B)= \frac{1}{n} {\textstyle \sum_{i}^{n}} \left \|  A_{i} - B_{i}\right \| _{2}$ is the matching cost function.
$\rho$ is the tensor rolling function, and we use code expression for clarity:
\begin{equation}
\begin{split}
    \rho(\mathcal{D}, \varphi,\mathrm{dim})& = ( \left \lceil \varphi \right \rceil  - \varphi)\mathrm{torch.roll}(\mathcal{D}, \left \lfloor \varphi \right \rfloor, \mathrm{dim}) \\
    & + (\varphi - \left \lfloor \varphi \right \rfloor )\mathrm{torch.roll}(\mathcal{D}, \left \lceil \varphi \right \rceil, \mathrm{dim}).
\end{split}
\end{equation}
This rolling function is designed to handle cases where the optimal value of $\varphi$ is not an integer, allowing interpolation between discrete shifts of $\mathcal{D}$ along the angle dimension.

\begin{algorithm}[h]
    \caption{BEV Padding}
    \label{alg:pad}
    \KwIn{$Q \in \mathbb{R}^{H \times W \times C}$, $t = (x, y)$, grid size $g$}
    \KwOut{$Q^{t} \in \mathbb{R}^{H \times W \times C}$}
    
    $Q^{t} \gets \rho(\rho(Q, \tfrac{-x}{g}, 0), \tfrac{-y}{g}, 1)$\;
    $Q^{t}[H - \lceil \tfrac{x}{g} \rceil:, :, :] \gets 1 \;\;\mathbf{if}\; x \geq 0 \;\mathbf{else}\; Q^{t}[:\lceil \tfrac{x}{g} \rceil, :, :] \gets 1$\;
    $Q^{t}[:, W - \lceil \tfrac{y}{g} \rceil:, :] \gets 1 \;\;\mathbf{if}\; y \geq 0 \;\mathbf{else}\; Q^{t}[:, :\lceil \tfrac{y}{g} \rceil, :] \gets 1$\;
    \Return $Q^{t}$
\end{algorithm}

\noindent{\bf Metric Localizer}
Note that the yaw estimator inherently neglects the influence of translation between the two frames. 
To improve this, we propose the metric localizer for complete pose estimation.
Initially, we hypothesize that the translation between two frames is $\hat{t}_{s\overset{}{\rightarrow}t }$ within a specific spatial domain.
Then, we apply BEV Padding (Alg.\ref{alg:pad}) to the source BEV queries $Q_{s}$ using $\hat{t}$.
The underlying rationale is that when the ground-truth translation is used to pad the source BEV queries, followed by yaw estimation, the resulting yaw should minimize the matching cost between the padded source BEV queries and the target BEV queries. 
Thus, pose estimation becomes a search for the translation and yaw that yield the lowest matching cost. 
In practice, we employ a coarse-to-fine strategy: first, a coarse grid search narrows the range of possible translations, followed by random sampling within the refined range to complete the fine search.

\noindent{\bf Re-ranking Strategy}
\label{subsec: rr}
We designe a re-ranking strategy based on the metric localizer for global localization. 
For each map candidate in $\mathcal{C}$ get by topological localizer, we first perform yaw estimation on it with the source frame, but without applying BEV Padding. 
The map candidates are then re-ranked according to the matching costs. 
The \textit{Top-1} map frame, identified as the one with the lowest matching cost, is considered the closest to the source frame. 
Subsequently, we employ the complete metric localizer to estimate the relative pose between the selected map frame and the source frame.

\subsection{Loss Functions}
We set triplets $[I_{s},I_{p},I_{n}]$ for model learning, where $I_{s}$ is the source sample, $I_{p}$ is a positive sample within 2\textit{m} of $I_{s}$, and $I_{m}$ is a negative sample more than 3\textit{m} away from $I_{s}$.

\noindent{\bf Triplet Margin Loss}
We adopt the standard triplet margin loss for global descriptor learning: 
\begin{equation}
    \mathcal{L}_{trp}(\mathcal{G}_{s}, \mathcal{G}_{p}, \mathcal{G}_{n}) = \max(\left \|  \mathcal{G}_{s}  - \mathcal{G}_{p} \right \|_{2} - \left \|  \mathcal{G}_{s}  - \mathcal{G}_{n}\right \|_{2} + \vartheta_{trp}, 0),
\end{equation}
Here, $\mathcal{G}_{s}$, $\mathcal{G}_{p}$, $\mathcal{G}_{n}$ are the global descriptor of the source, positive, negative sample.
$\vartheta_{trip}=0.1$ is the triplet margin.

\noindent{\bf Hard Sample Mining}
As training progresses, many triplets no longer contribute to the loss, reducing efficiency. 
To address this, we set a hard sample mining strategy that resamples triplets the model finds difficult until they are consistently classified correctly.

\noindent{\bf Hinge Loss}
For loop closure detection, we additionally add hinge loss to constrain the distance between global descriptors:
\begin{equation}
\begin{split}
    \mathcal{L}_{hinge}(\mathcal{G}_{s}, \mathcal{G}_{p}, \mathcal{G}_{n}) = & \max(\left \|  \mathcal{G}_{s}- \mathcal{G}_{p} \right \|_{2} - \vartheta_{p}^{h}, 0) + \\ & \max(\vartheta_{n}^{h} - \left \|  \mathcal{G}_{s}- \mathcal{G}_{n} \right \|_{2} , 0) ,
\end{split}
\end{equation}
where $ \vartheta_{p}^{h}=0.1$ and $\vartheta_{p}^{n}=0.2$ are the positive and negative global descriptor margin respectively. 

\noindent{\bf Circle-based Yaw Matching Loss}
The local-view descriptor learning is based on circle loss\cite{sun2020circle}.
We first pad the source BEV queries with ground-truth translation $Q^{t_{s\rightarrow p}}_{s} =\mathrm{BEVPad}(Q_{s}, t_{s\rightarrow p})$, and then get the local-view descriptor $\mathcal{D}^{t_{s\rightarrow p}}_{s}$ from it.
Finally, we roll the local-view descriptor with the ground-truth yaw $\varphi_{s\rightarrow p}$ to get the source local-view descriptor that geometrically aligned with the positive local-view descriptor $\mathcal{D}_{s}^{\xi _{s\rightarrow p}} = \rho(\mathcal{D}_{s}^{t_{s\rightarrow p}}, -T\cdot\frac{\varphi_{s\rightarrow p}}{360}, 0)$.
The loss is calculated between $\mathcal{D}_{s}^{\xi _{s\rightarrow p}}$ and $\mathcal{D}_{p}$ in circle loss style:
\begin{footnotesize}
\begin{equation}
\begin{split}
    \mathcal{L}_{yaw}(\mathcal{D}_{s}^{\xi _{s\rightarrow p}}, \mathcal{D}_{p}) = \frac{1}{T} \sum_{0\le i <T}^{} & \log(1 + e^{\gamma(\delta(\mathcal{D}_{s, i}^{\xi _{s\rightarrow p}}, \mathcal{D}_{p,i}) - \vartheta_{p}^{y})^{2}}\cdot \\
    & \sum_{\left | j-i \right | > 0}^{}e^{\gamma(\vartheta_{n}^{y} - \delta(\mathcal{D}_{s, i}^{\xi _{s\rightarrow p}}, \mathcal{D}_{p,j}))^{2}} ),
\end{split}
\end{equation}
\end{footnotesize}
where $\gamma=40$ is the log scale, $\vartheta_{p}^{y}=0.1$ and $\vartheta_{n}^{y}=1.4$ are the positive and negative local-view descriptor margin respectively.
Such learning ensures that each source local-view descriptor is only similar to its own geometrically aligned positive local-view descriptor, so that the yaw estimator can solve the optimal result.

\noindent{\bf Translation Bias Loss}
Near the ground true pose $\xi_{s \rightarrow p}$, we first randomly sample a biased pose $\tilde{\xi}_{s \rightarrow p}=(x_{s \rightarrow p} + \Delta_{x}, y_{s \rightarrow p} + \Delta_{y}, \varphi_{s \rightarrow p})$.
Then, we get the source local-view descriptor $\mathcal{D}_{s}^{\tilde{\xi} _{s\rightarrow p}}$ with the biased pose.
The translation bias loss is computed as:
\begin{equation}
\begin{split}
    \mathcal{L}_{tra}&(\mathcal{D}_{s}^{\xi _{s\rightarrow p}}, \mathcal{D}_{s}^{\tilde{\xi} _{s\rightarrow p}},\mathcal{D}_{p}) = \\
    & \max(\delta(\mathcal{D}_{s}^{\xi_{s\rightarrow p}}, \mathcal{D}_{p}) - \delta(\mathcal{D}_{s}^{\tilde{\xi} _{s\rightarrow p}}, \mathcal{D}_{p}) + \vartheta_{tra}, 0),
\end{split}
\end{equation}
where $\vartheta_{tra}=0.1$ is the bias margin.
Such learning ensures that the closer to ground-truth translation for BEV Padding, the lower cost produced when doing yaw estimation, so that the metric localizer can find the optimal translation.

\definecolor{Silver}{rgb}{0.752,0.752,0.752}
\begin{table*}
\caption{\textbf{Quantitative Global Localization Results on the NCLT and the Oxford Dataset.}}
\centering
\scalebox{0.78}{
\begin{tabular}{l|ll|ll|ll|ll} 
\hline
\multirow{2}{*}{\textbf{Methods}} & \multicolumn{4}{c|}{\textbf{NCLT}}                                                                                                                                                                                                        & \multicolumn{4}{c}{\textbf{Oxford}}                                                                                     \\ 
\cline{2-9}
                                  & \textbf{Recall@1}$\uparrow$                                 & \textbf{Recall@5}$\uparrow$                                 & \textbf{AOE/APE@25\%}$\downarrow$                   & \textbf{AOE/APE@50\%}$\downarrow$                   & \textbf{Recall@1}$\uparrow$ & \textbf{Recall@5}$\uparrow$ & \textbf{AOE/APE@25\%}$\downarrow$ & \textbf{AOE/APE@50\%}$\downarrow$  \\ 
\hline
NetVLAD$^{\star, \dagger}$\cite{arandjelovic2016netvlad}                           & 52.6\%                                                       & 71.0\%                                                       & -/-                                                  & -/-                                                  & 76.6\%                       & \textbf{91.3}\%                       & -/-                                & -/-                                 \\
How-ASMK$^{\star}$\cite{tolias2020learning}                      & 55.0\%                                                       & 76.3\%                                                       & -/-                                                  & -/-                                                  & 80.0\%                       & 90.3\%                       & -/-                                & -/-                                 \\
DOLG$^{\star}$\cite{yang2021dolg}                              & 54.5\%                                                       & 75.3\%                                                       & -/-                                                  & -/-                                                  & 62.0\%                       & 82.5\%                       & -/-                                & -/-                                 \\
vDISCO\cite{xu2023leveraging}                            & 74.5\%                                                       & 84.9\%                                                       & 1.8/-                                                & 3.9/-                                                & 79.0\%                       & 88.8\%                       &0.8/-                                    &2.1/-                                     \\ 
\hline
\textbf{VVLoc-G}                         & 25.5\%                                                       & 42.2\%                                                       & -/-                                                  & -/-                                                  &28.4\%                       &44.8\%                      & -/-                                & -/-                                 \\
\textbf{VVLoc-G-ML}                      & {\cellcolor[rgb]{0.753,0.753,0.753}}62.4\% (+36.9\%)         & {\cellcolor[rgb]{0.753,0.753,0.753}}75.3\% (+33.1\%)         & {\cellcolor[rgb]{0.753,0.753,0.753}}1.0/0.4          & {\cellcolor[rgb]{0.753,0.753,0.753}}2.5/1.4          &{\cellcolor[rgb]{0.753,0.753,0.753}}59.5\%(+31.1\%)               &{\cellcolor[rgb]{0.753,0.753,0.753}}66.6\%(+21.8\%)                &{\cellcolor[rgb]{0.753,0.753,0.753}}0.4/0.6                            &{\cellcolor[rgb]{0.753,0.753,0.753}}1.0/1.4                             \\ 
\hline
\textbf{VVLoc-N}                         & 35.4\%                                                       & 53.3\%                                                       & -/-                                                  & -/-                                                  &62.5\%                               & 76.0\%                             & -/-                                & -/-                                 \\
\textbf{VVLoc-N-ML}                      & {\cellcolor[rgb]{0.753,0.753,0.753}}59.5\% (+24.1\%)         & {\cellcolor[rgb]{0.753,0.753,0.753}}66.7\% (+13.4\%)         & {\cellcolor[rgb]{0.753,0.753,0.753}}1.1/0.6          & {\cellcolor[rgb]{0.753,0.753,0.753}}2.8/1.3          &{\cellcolor[rgb]{0.753,0.753,0.753}}77.8\% (+15.3\%)                             &{\cellcolor[rgb]{0.753,0.753,0.753}}85.0\% (+9.0\%)                             &{\cellcolor[rgb]{0.753,0.753,0.753}}0.4/0.6                                   &{\cellcolor[rgb]{0.753,0.753,0.753}}0.8/1.0                                    \\ 
\hline
\textbf{VVLoc-V}                         & 75.9\%                                                       & 86.2\%                                                       & -/-                                                  & -/-                                                  &76.2\%                              &87.5\%                              & -/-                                & -/-                                 \\
\textbf{VVLoc-V-ML}                      & {\cellcolor[rgb]{0.753,0.753,0.753}}\textbf{80.6\% }(+4.7\%) & {\cellcolor[rgb]{0.753,0.753,0.753}}\textbf{89.7\%} (+3.5\%) & {\cellcolor[rgb]{0.753,0.753,0.753}}\textbf{0.7/0.3} & {\cellcolor[rgb]{0.753,0.753,0.753}}\textbf{1.6/0.7} &{\cellcolor[rgb]{0.753,0.753,0.753}}\textbf{83.7\%} (+7.5\%)                             &{\cellcolor[rgb]{0.753,0.753,0.753}}90.9\% (+3.4\%)                             &{\cellcolor[rgb]{0.753,0.753,0.753}}\textbf{0.3}/\textbf{0.5}                                   &{\cellcolor[rgb]{0.753,0.753,0.753}}\textbf{0.7}/\textbf{0.9}                                    \\
\hline
\end{tabular}
}
\label{tab:global}
\begin{scriptsize}
\begin{tablenotes}
  \item[1] 
  ${}^{\star}$ Using panorama images as input; ${}^{\dagger}$ Pretrained on the Pittsburgh dataset; \textbf{-}: Not applicable
\end{tablenotes}
\end{scriptsize}

\end{table*}

\begin{figure*}[htb]
  \centering
  \includegraphics[width=1.0\linewidth]{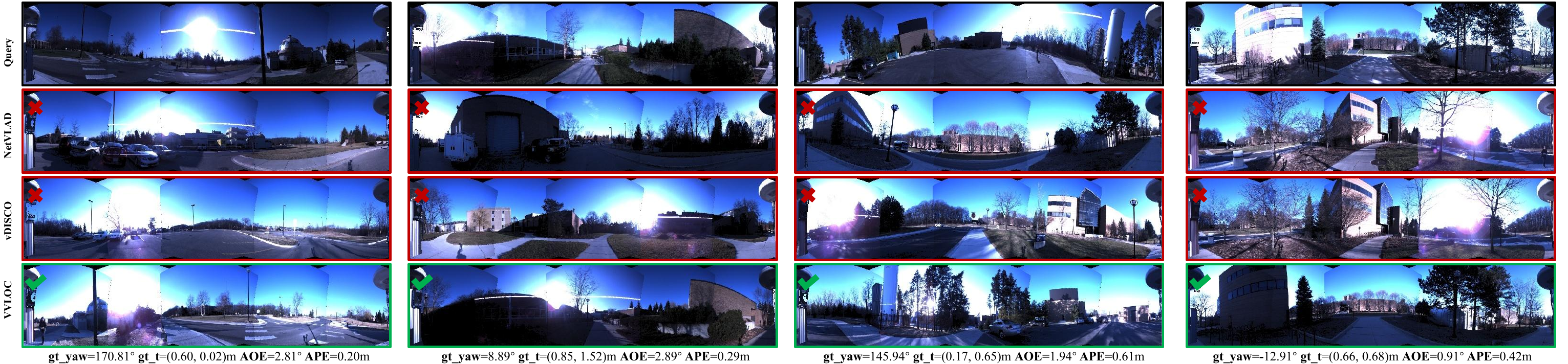}
   \caption{\textbf{Qualitative Global Localization Results on the NCLT Dataset.} VVLoc can retrieve the correct map frames with a wide range of perspectives and lighting changes, and give correct metric pose estimation.}
   \label{fig:nclt}
\vspace{-1em}
\end{figure*}

\begin{figure}[htb]
  \centering
  \includegraphics[width=1.0\linewidth]{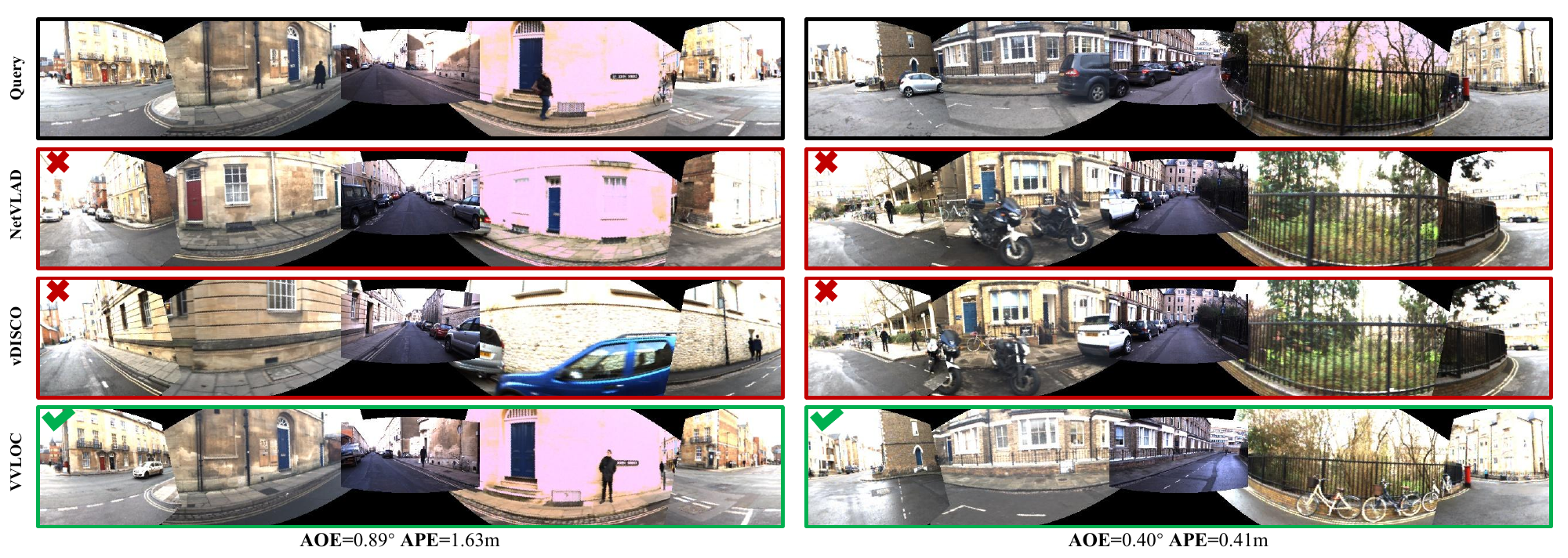}
   \caption{\textbf{Qualitative Global Localization Results on the Oxford Radar Robotcar Dataset.} }
   \label{fig:oxford}
\vspace{-2em}
\end{figure}

\section{Experiments}
\subsection{Datasets and Model Details}
Experiments were done on two public and one self-collected datasets.

\noindent{\bf NCLT Dataset\cite{carlevaris2016university}} is a large-scale and long-term dataset collected on the University of Michigan's North Campus by a Segway robot.
The NCLT spans 15 months and includes a wide range of environmental changes, such as moving individuals, seasonal changes, and structural changes.

\noindent{\bf Oxford Radar Robotcar (Oxford) Dataset\cite{RobotCarDatasetIJRR}} is an addition to the Oxford RobotCar Dataset for the study of autonomous driving.
In January 2019, 32 traversals of a central Oxford route were recorded. 
This dataset includes a wide range of weather and lighting situations.

\noindent{\bf Park-mapping Dataset} is a large collection of data from various parking areas, including ground level lots, multilayer underground lots, and service areas. 
It covers diverse conditions such as seasonal and lighting changes, environmental degradation, and repetitive patterns, making it ideal for testing localization algorithm.
The data collection vehicle was equipped with a hybrid mechanical LiDAR (FoV 120° $\times$ 25°), two front and rear pinhole cameras, and two side fisheye cameras. 
All sensors were well calibrated, and images were undistorted before use.
Raw poses were provided by a Novatel 100\textit{c} optical fiber positioning system (2\textit{cm} / 0.01° precision at 100\textit{Hz}).
We keyframed the data based on 2\textit{m} distance or 5° rotation, refined the poses using LiDAR odometry, and applied global pose graph optimization to obtain accurate, densely sampled poses strictly aligned with camera images as the training ground-truth.
Data from 280 places were used for training, and evaluation was performed on data from an additional 44 places.

\noindent{\bf Configurations}
We used PyTorch framework and 8 A-100 GPUs for training.
We used the same hyperparameters for all datasets to ensure model versatility.
Images were resized to 256$\times$448.
Inference between two frames used 5471\textit{MB} of CUDA memory.
The training batch size was set to 4, and the model was trained in two stages: 20 epochs for global descriptor training, followed by 20 epochs for joint global-local-view descriptor training. 
Both stages used the ADAM optimizer with an initial learning rate of $1e^{-4}$ and a decay rate of 0.95 per epoch.
For the \textit{Cartesian} BEV, we set the BEV size to 100 $\times$ 100, grid size $g = 0.3m$, and feature dimension $C = 256$.
For the polar BEV, $T$ and $R$ were set to 120 and 40 respectively.
The temperature factor, $\sigma_{r}$ and $\sigma_{a}$ were set to 4.8 and 15 respectively.
For global localization, we retrieved top 20 map candidates for re-ranking.
We consider matching cost as the confidence.

\noindent{\bf Runtime}
VVLoc takes 22.36\textit{ms} for image feature extraction, 58.78\textit{ms} for BEV feature extraction, 5.24\textit{ms} for global descriptor extraction, 11.40\textit{ms} for local-view descriptor extraction.
Once features are extracted, VVLoc takes 0.09\textit{ms}  for topological localization, 238\textit{ms} for metric localization, and 261\textit{ms} for re-ranking.
All of the data above are tested under A-100 GPUs with an AMD EPYC 7713 64-Core CPU.

\subsection{Global Localization Evaluation}
We first performed a global localization evaluation on two public datasets, with experimental settings strictly aligned with previous work\cite{xu2023leveraging}.
VVLoc was tested with various configurations: \textbf{G}, \textbf{N}, and \textbf{V} denote the use of GeM, NetVLAD, and vDISCO pooling, respectively, for generating global descriptors, while \textbf{ML} represents the inclusion of the metric localizer for re-ranking and pose estimation. 

Comparison results with the state-of-the-art baselines are presented in Tab.\ref{tab:global}, Fig.\ref{fig:nclt}, and Fig.~\ref{fig:oxford}.
VVLoc achieves superior performance on the NCLT dataset, significantly surpassing baselines across all metrics. 
Its results on the Oxford dataset are slightly less competitive, with a marginally higher position error, likely due to poor calibration and synchronization in the Oxford dataset, which hinders the training of the geometry-dependent VVLoc. 
The proposed metric localizer is a key component, enhancing topological localization accuracy and enabling precise 3-DoF pose estimation, a capability not achievable with prior methods.

\begin{table}
\centering
\caption{\textbf{Quantitative Loop Closure Detection Results on the Park-mapping Dataset.}}
\label{tab:lcd}
\scalebox{0.73}{
\begin{tabular}{l|lll|ll} 
\hline
\multirow{2}{*}{\textbf{Methods}} & \multicolumn{3}{c}{\textbf{\textbf{Underground}}}                                                     & \multicolumn{2}{c}{\textbf{\textbf{Outdoor}}}                                                                      \\ 
\cline{2-6}
                                  & \textbf{Precision@2m}$\uparrow$ & \textbf{Recall@2m}$\uparrow$ & \textbf{CFER}$\downarrow$ & \textbf{\textbf{Precision@2m}$\uparrow$} & \textbf{\textbf{Recall@2m}$\uparrow$}              \\ 
\hline
BEV-CV~\cite{shore2023bev}                           & 58.05\%                      & 66.73\%                      & 1.13\%                    & 62.90\%                                & 67.28\%                                          \\
vDISCO~\cite{xu2023leveraging}                           & 68.77\%                      & \textbf{68.51\%}                      & 0.16\%                        & 67.43\%                               & 68.69\%                                          \\
\textbf{VVLoc-G}                          & {\cellcolor[rgb]{0.753,0.753,0.753}}65.13\%                 & {\cellcolor[rgb]{0.753,0.753,0.753}}68.11\%                 & {\cellcolor[rgb]{0.753,0.753,0.753}}0.26\%                    & {\cellcolor[rgb]{0.753,0.753,0.753}}65.15\%                          & {\cellcolor[rgb]{0.753,0.753,0.753}}\textbf{68.75\%}                                      \\
\textbf{VVLoc-V}                           & {\cellcolor[rgb]{0.753,0.753,0.753}}\textbf{73.39\%}                 & {\cellcolor[rgb]{0.753,0.753,0.753}}68.44\%              & {\cellcolor[rgb]{0.753,0.753,0.753}}\textbf{0.01\%}                    & {\cellcolor[rgb]{0.753,0.753,0.753}}\textbf{71.99\%}                          & {\cellcolor[rgb]{0.753,0.753,0.753}}68.08\%  \\
\hline
\end{tabular}
}
\vspace{-1em}
\end{table}

\begin{figure*}
  \includegraphics[width=0.98\textwidth]{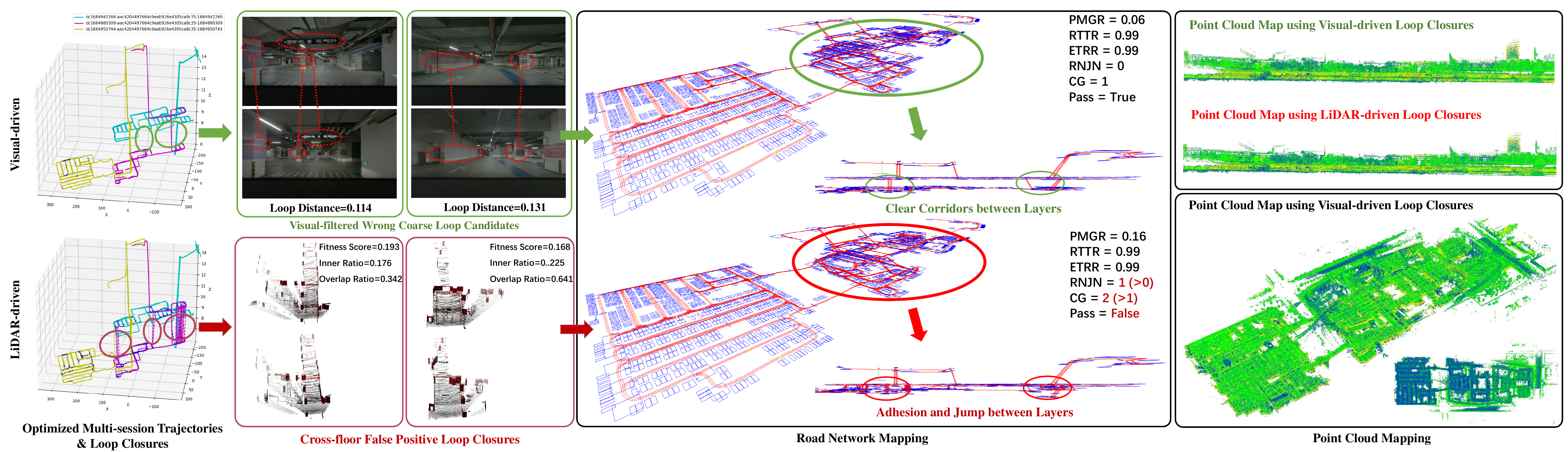}
  \caption{\textbf{Qualitative Multi-modal Mapping Evaluation.} We perform road network and point cloud mapping using loop closures from different localization methods. The proposed VVLoc filters false loops via image textures, whereas the LiDAR-driven method fails against structurally similar false positives, leading to mapping failure in the multi-layer underground parking lot.}
  \label{fig:map}
\vspace{-1em}
\end{figure*}

\begin{table}
\caption{\textbf{Quantitative Metric Pose Estimation Results of Detected Loop Pairs on the Park-mapping Dataset.}} 
\centering
\scalebox{0.85}{
\begin{tabular}{l|llll} 
\hline
\textbf{Underground}           & \textbf{mRRE} $\downarrow$                     & \textbf{mRTE}$\downarrow$                     & \textbf{RR(\textless{}5deg)}$\uparrow$             & \textbf{RR(\textless{}2m})$\uparrow$    \\ 
\hline
Geotransformer\cite{qin2023geotransformer}              & 30.66                                     & 1.79                                     & 43.49\%   &    67.30\%                                          \\
BEV-Locator\cite{zhang2022bev}                     & 15.31                                     & 1.32                                     &       76.01\%                                      &    83.92\%                                          \\
\textbf{VVLoc-V-ML}                         & {\cellcolor[rgb]{0.753,0.753,0.753}}10.07 & {\cellcolor[rgb]{0.753,0.753,0.753}}0.68 & {\cellcolor[rgb]{0.753,0.753,0.753}}82.89\% & {\cellcolor[rgb]{0.753,0.753,0.753}}94.74\%  \\
\textbf{VVLoc-V-ML}$^{\star}$                         & {\cellcolor[rgb]{0.753,0.753,0.753}}\textbf{3.18}  & {\cellcolor[rgb]{0.753,0.753,0.753}}\textbf{0.51} & {\cellcolor[rgb]{0.753,0.753,0.753}}\textbf{93.21\%} & {\cellcolor[rgb]{0.753,0.753,0.753}}\textbf{98.81\%}  \\ 
\hline
\textbf{Outdoor} & \textbf{mRRE}$\downarrow$              & \textbf{mRTE}$\downarrow$             & \textbf{RR(\textless{}5deg)}$\uparrow$     & \textbf{RR(\textless{}2m})$\uparrow$        \\ 
\hline
Geotransformer\cite{qin2023geotransformer}                 & 22.45                                     & 1.64                       &  51.72\%                                     &      74.12\%                                 \\
BEV-Locator\cite{zhang2022bev}                     & 13.51                                     & 0.96                                     &      82.43\%                      &    85.95\%                                    \\
\textbf{VVLoc-V-ML}                         & {\cellcolor[rgb]{0.753,0.753,0.753}}3.53  & {\cellcolor[rgb]{0.753,0.753,0.753}}0.63 & {\cellcolor[rgb]{0.753,0.753,0.753}}92.34\% & {\cellcolor[rgb]{0.753,0.753,0.753}}97.70\%  \\
\textbf{VVLoc-V-ML}$^{\star}$                          & {\cellcolor[rgb]{0.753,0.753,0.753}}\textbf{2.04}  & {\cellcolor[rgb]{0.753,0.753,0.753}}\textbf{0.57} & {\cellcolor[rgb]{0.753,0.753,0.753}}\textbf{96.04\%} & {\cellcolor[rgb]{0.753,0.753,0.753}}\textbf{98.23\%}  \\
\hline
\end{tabular}
}
\begin{scriptsize}
\begin{tablenotes}
  \item[1] 
  ${}^{\star}$ Results with matching cost lower than 0.75.
\end{tablenotes}
\end{scriptsize}
\label{tab:mpe}
\vspace{-1em}
\end{table}

\subsection{Loop Closure Detection Evaluation}
To evaluate localization in more complex scenarios, we conducted loop detection evaluation on the Park-mapping dataset.
Each map in the dataset includes multi-session trajectories from the same parking place. 
Each session contains consecutive keyframes, with each keyframe consisting of multi-camera images, LiDAR point cloud, and initial global pose measurements. 
Using VVLoc's topological localizer, we identified loop closures among all keyframes, both within and across sessions. 
VVLoc's metric localizer was then applied to estimate the relative pose for all detected loop closures.

For baselines, we trained vDISCO~\cite{xu2023leveraging} and BEV-CV~\cite{shore2023bev} for loop closure detection comparison, where BEV-CV is reproduced without map segmentation supervision. 
Both baselines are trained with an additional hinge loss.
For 3-DoF pose estimation between the detected loop closures, we reproduced two baselines for comparison:
\textbf{(i) Regression-based Approach}.
BEV-locator\cite{zhang2022bev} proposed metric localization from visual BEV to semantic vector maps using a Transformer encoder–decoder to regress 3-DoF poses. 
We reproduced this method and adapted it by replacing map features with another visual BEV to predict relative poses between two observations.
\textbf{(ii) Registration-based Approach}.
We treat the visual BEV as a point cloud with explicit 3D coordinates and features in the ego-vehicle system, training Geotransformer\cite{qin2023geotransformer}, a leading point cloud registration method, to estimate metric relative poses between frames.

For evaluation, we used 
\textbf{(i) Recall@2m} and \textbf{(ii) Precision@2m} to assess recall rate and precision of loop closure detection with 2\textit{m} threshold.
\textbf{(iii) Cross-floor Error Rate (CFER)} to quantify false closures across different floors in multi-level environments.
\textbf{(iv) mean Relative Rotation Error (mRRE)} and \textbf{(v) mean Relative Translation Error (mRTE)} to measure pose accuracy in terms of rotational and translational deviations.
\textbf{(iv) Registration Recall (RR)} to evaluate the proportion of pairs with successfully estimated relative poses under predefined error thresholds.

Tab.\ref{tab:lcd} presents the loop closure detection results in 9 places.
Given its higher performance and lower cross-floor error rate, we further conduct metric pose estimation evaluation based on VVLoc-V's loop detection results.
The pose estimation results (Tab.\ref{tab:mpe}) show that point cloud registration and regression-based methods suffer from high errors and poor false positive filtering, with the former's limitations aligning with our analysis in Fig.~\ref{fig:idea}. 
In contrast, VVLoc achieves superior pose estimation accuracy using a novel matching-based strategy, effectively filtering erroneous loops, particularly in complex underground places, demonstrating its reliability and practicality for real-world applications.

\subsection{Multi-modal Mapping Evaluation}
While existing localization methods are seldom evaluated on downstream tasks, particularly in terms of their impact on mapping quality, we bridge this gap by integrating VVLoc into our self-developed multi-modal mapping frameworks, including both point cloud and road network mapping. 
Given ongoing interest in vision–LiDAR trade-offs, we direct validated our proposed VVLoc in practical mapping scenarios through comprehensive comparisons with a state-of-the-art LiDAR-based loop closure baseline.
The loop pair extraction process differs slightly from the previous experiment: we first identified all keyframes within 20\textit{m} of a keyframe's initial position measurement as coarse loop pairs. 
Then, we use the following approaches to refine valid loop pairs:
\textbf{(i) Visual-driven Approach}.
We used VVLoc’s topological localizer to filter the coarse loop pairs, keeping those with a global descriptor distance under 0.1. 
Then, we applied its metric localizer combined with ICP\cite{arun1987least} to estimate the relative poses. 
Loop closures with a matching cost under 0.75 and a fitness score under 0.2 are considered valid;
\textbf{(ii) LiDAR-driven Approach}.
We employed Predator\cite{huang2021predator}, a leading point cloud registration method finely trained on our data, combined with Sc$^{2}$-pcr\cite{chen2022sc2}, for coarse registration. 
Fine registration was then performed using ICP\cite{arun1987least}. 
Valid loop closures have a fitness score under 0.2, an overlap ratio exceeding 0.3, and an inner ratio exceeding 0.1.

The loop closures were integrated into the mapping process to optimize the map, and map quality was used as a quantitative evaluation.
For evaluation metrics, we used \textbf{(i) Point Mapping Ghost Rate~\cite{chen20203d} (PMGR)} to measure the degree of misalignment or duplication in the mapped point cloud. 
\textbf{(ii) Road to Trajectory Rate (RTTR)} to measure the coverage completeness of the mapped road network relative to the collected trajectories.
\textbf{(iii) Enter to Road Rate (ETRR)} to measure the success rate of routing from road network entry points to other roads.
\textbf{(iv) Road Network Jumping Number (RNJN)} to measure the number of discontinuities or sudden changes in the road network.
\textbf{(v) Connected Graphs (CG)} to count the number of connected graphs in the mapped road network.
\textbf{(vi) Pass Rate (PR)} to measure the proportion of evaluation places where the mapping quality meets the threshold (\textbf{RTTR}$>$0.7, \textbf{ETRR}$>$0.75, \textbf{RNJN}$=$0, \textbf{CG}$=$1).

\definecolor{Silver}{rgb}{0.752,0.752,0.752}
\begin{table}
\caption{\textbf{Quantitative Point Cloud and Road Network Mapping Results on the Park-mapping Dataset.}}
\centering
\scalebox{0.82}{
\begin{tblr}{
  cell{3}{2} = {Silver},
  cell{3}{3} = {Silver},
  cell{3}{4} = {Silver},
  cell{3}{5} = {Silver},
  cell{3}{6} = {Silver},
  cell{3}{7} = {Silver},
  vline{2} = {-}{},
  hline{1-2,4} = {-}{},
}
\textbf{Methods} & \textbf{PMGR}$\downarrow$ & \textbf{RTTR}$\uparrow$ & \textbf{ETRR}$\uparrow$ & \textbf{RNJN}$\downarrow$ & \textbf{CG}$\downarrow$ & \textbf{PR}$\uparrow$ \\
LiDAR\cite{huang2021predator, chen2022sc2}     &    0.293           &     0.985          &  0.904             &   0.286            &  1.257           &   68.57\%          \\
\textbf{Visual (Ours)}   &       \textbf{0.271}        &    \textbf{0.986}           &       \textbf{0.920}        &      \textbf{0.257}         &     \textbf{1.086}        &  \textbf{80.00}\%           
\end{tblr}
}
\label{tab:mapping}
\vspace{-1em}
\end{table}

\begin{figure}[htb]
  \centering
  \includegraphics[width=1.0\linewidth]{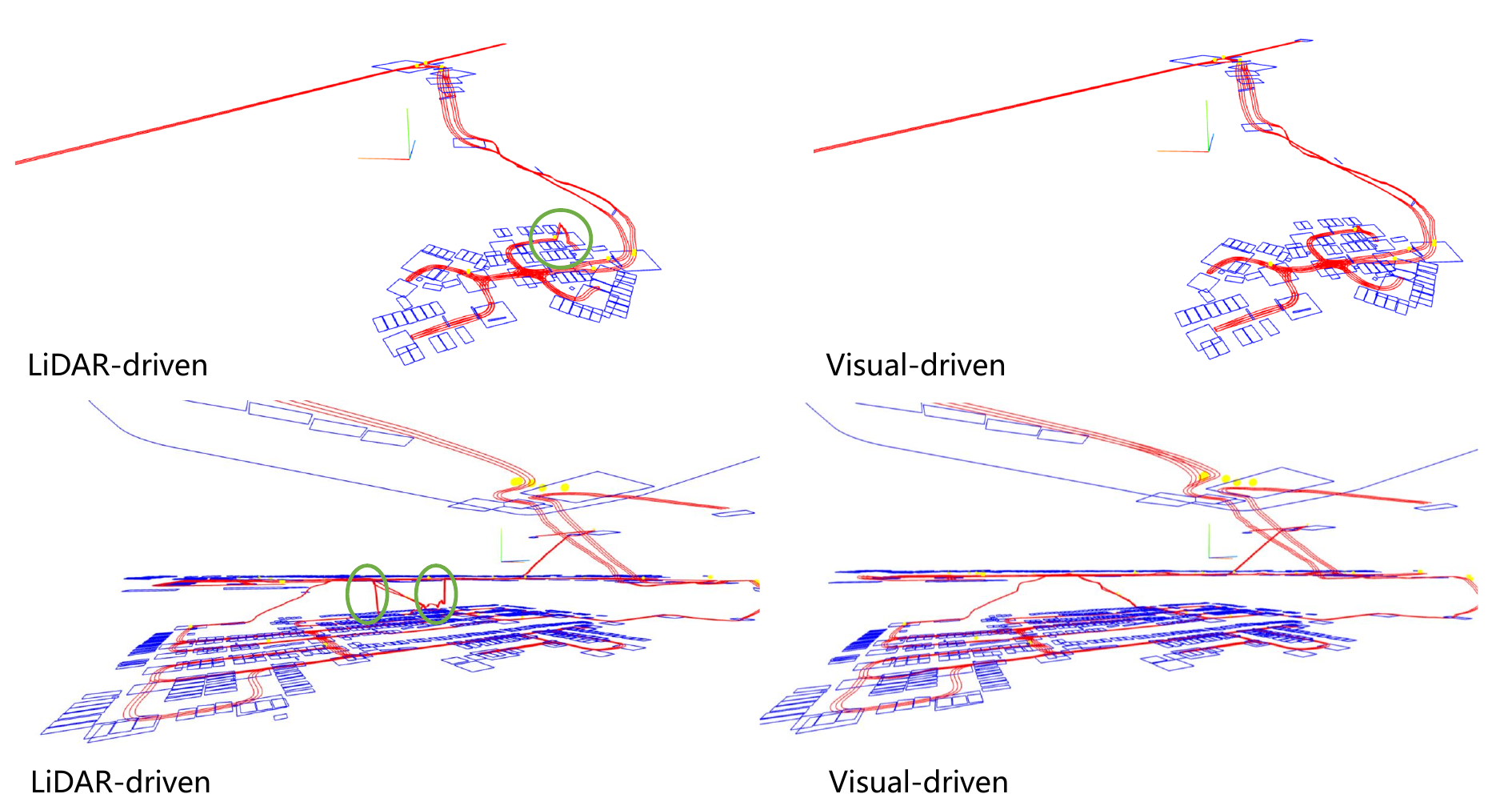}
   \caption{\textbf{Additional Qualitative Mapping Results.} The green circles represent the road network jumps that make the results not passed.}
   \label{fig:map_supp}
\vspace{-1em}
\end{figure}

Tab.\ref{tab:mapping} summarizes the mapping performance in 35 places. 
Compared to the LiDAR-driven approach, our visual-driven approach demonstrates superior performance in both point cloud mapping and road network mapping. 
For instance, point cloud maps generated using visual loop closures exhibit a lower ghosting rate, while road network maps achieve higher connectivity and reduced jump rates.
Fig.\ref{fig:map} illustrates the mapping results in a multi-layer underground parking lot. 
The LiDAR-driven approach struggles with structural similarities between levels, causing false loop closures that lead to mapping errors such as road network sticking and jumping. 
In contrast, the visual-driven method leverages subtle textures and patterns, accurately separating corridors between layers and delivering significantly better quality.
More representative qualitative results are shown in Fig~\ref{fig:map_supp}.

\subsection{Localization Evaluation in Degraded Scenarios}
\begin{figure}
    \centering
    \includegraphics[width=1.0\linewidth]{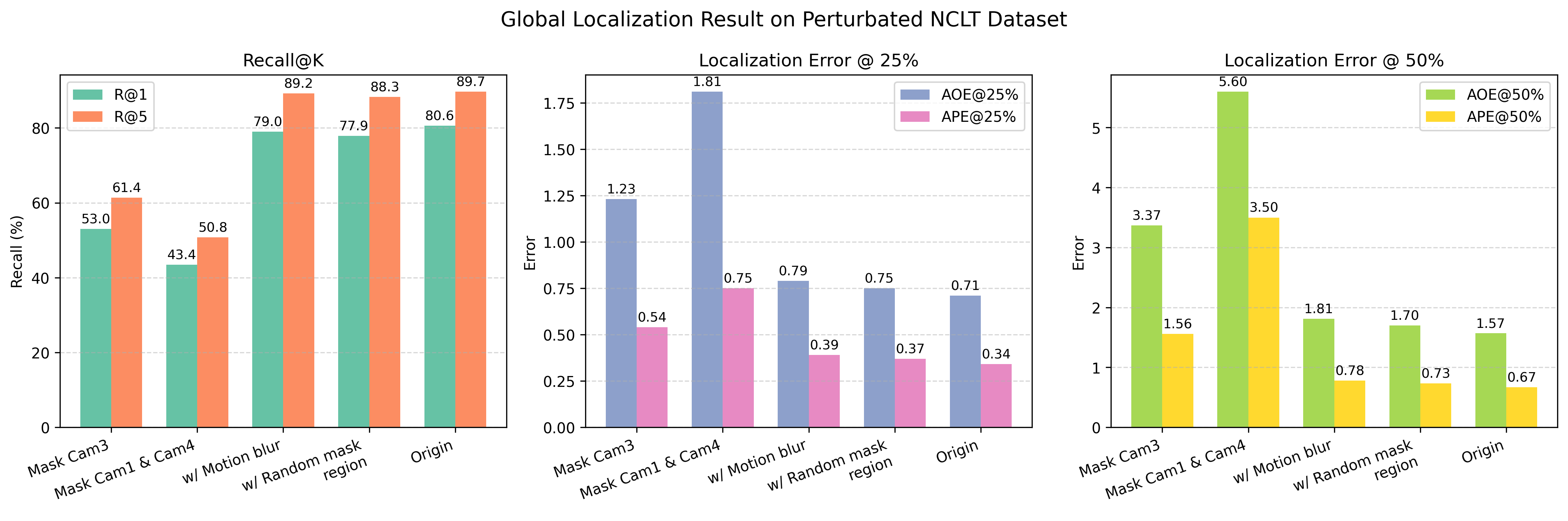}
    \caption{\textbf{Global Localization Results with Synthetic Perturbations on the NCLT Dataset.}}
    \label{fig:nclt_supp}
\vspace{-1em}
\end{figure}

\begin{figure}
    \centering
    \includegraphics[width=0.9\linewidth]{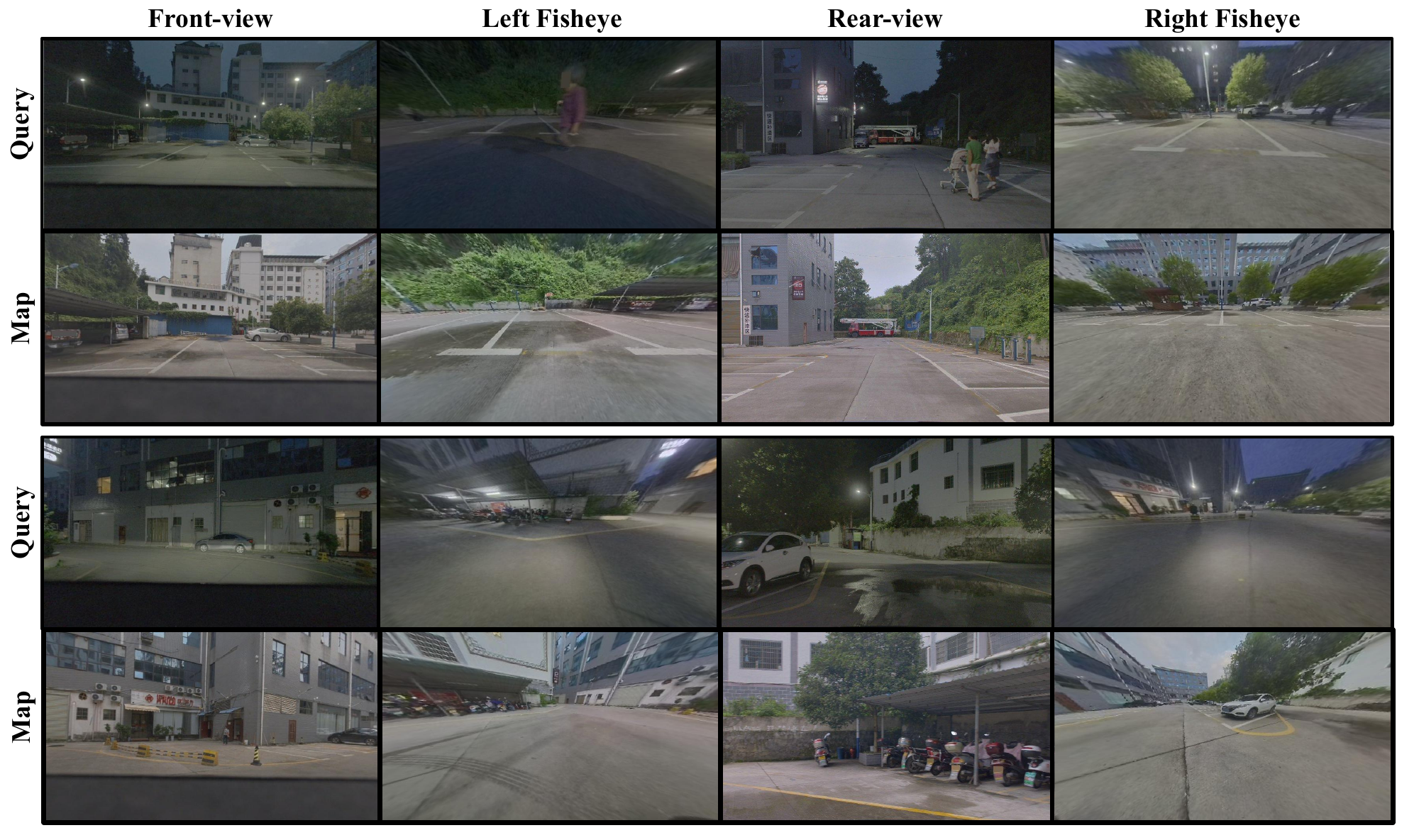}
    \caption{\textbf{Samples from Day-night Sequence.} VVLoc can retrieve and localize in both simple (top) and challenging (bottom) cases.}
    \label{fig:day-night}
\vspace{-1.5em}
\end{figure}

To evaluate VVLoc in extreme environments, we additionally collected or simulated a range of degraded scenarios. 
We then directly applied VVLoc to these inputs for localization without any fine-tuning on the perturbed data.
For the NCLT dataset, we used synthetic perturbations to the original test sequence by randomly masking cameras, adding motion blur, or masking image regions, with results shown in Fig.~\ref{fig:nclt_supp}.
For the Park-mapping dataset, we additionally collected a day-to-night sequence pair (samples in Fig.~\ref{fig:day-night}) for real-world global localization evaluation, with results shown in Tab.~\ref{tab:day-night}.
The results demonstrate that VVLoc maintains robust localization in these extreme scenarios without fine-tuning, highlighting its strong generalization ability and practical applicability.

\begin{table}
\caption{\textbf{Quantitative Global Localization Results on Day-Night Sequence.}} 
\centering
\scalebox{0.78}{
\begin{tabular}{l|llll} 
\hline
\textbf{Day-night} & \textbf{Recall@1}$\uparrow$              & \textbf{Recall@5}$\uparrow$             & \textbf{AOE/APE@25\%}$\downarrow$     & \textbf{AOE/APE@50\%}$\downarrow$        \\ 
\hline
\textbf{VVLoc-V}                         & {\cellcolor[rgb]{0.753,0.753,0.753}}26.96\%  & {\cellcolor[rgb]{0.753,0.753,0.753}}46.08\% & {\cellcolor[rgb]{0.753,0.753,0.753}}-/- & {\cellcolor[rgb]{0.753,0.753,0.753}}-/-  \\
\textbf{VVLoc-V-ML}                          & {\cellcolor[rgb]{0.753,0.753,0.753}}46.42\%  & {\cellcolor[rgb]{0.753,0.753,0.753}}59.73\% & {\cellcolor[rgb]{0.753,0.753,0.753}}0.22/0.47  & {\cellcolor[rgb]{0.753,0.753,0.753}}0.82/2.87  \\
\hline
\end{tabular}
}
\label{tab:day-night}
\vspace{-1em}
\end{table}

\begin{table}
\caption{\textbf{Additional Ablation Results on Outdoor Loop Pair Pose Estimation on the Park-mapping Dataset.} }
\centering
\scalebox{0.85}{
\begin{tblr}{
  vline{2,6} = {-}{},
  hline{1-2,8} = {-}{},
}
\textbf{ID} & \textbf{RASA} & \textbf{TASA} & \textbf{TBL} & \textbf{HVS} & \textbf{mRRE} $\downarrow$ & \textbf{mRTE} $\downarrow$ \\
0  &  $\surd$   &   $\surd$   &  $\surd$    &  $\surd$   & 3.53 & 0.63 \\
1  &$\surd$   &   $\surd$   &  $\surd$  &  GT   &   2.12        &  0.00 \\
2  &$\surd$   &   $\surd$   &  $\surd$  &     &    7.00       &  - \\
3  &    &   $\surd$  &   $\surd$  &  $\surd$   & 4.11 & 0.68 \\
4  &     &     &   $\surd$  &   $\surd$  &  5.27    &  0.84     \\
5  &     &     &     &   $\surd$  &     12.75      & 3.12              \\             
\end{tblr}
}
\label{tab:ab}
\vspace{-1em}
\end{table}

\subsection{Additional Ablation Studies}
Tab.\ref{tab:ab} shows the ablation results for key components and operations. 
The results clearly indicate that both $\mathrm{RASA}$ and $\mathrm{TASA}$ significantly enhance pose estimation accuracy (IDs 0, 3, and 4).
The absence of translation bias loss ($\mathrm{TBL}$) impairs the metric localizer's ability to find the optimal pose based on matching cost, resulting in a notable accuracy drop (IDs 4 and 5).
Additionally, without hypothesis verification-based searching ($\mathrm{HVS}$), translation estimation is hindered and rotation error increases by 3.47° (IDs 0 and 2). 
When the ground-truth translation is exerted for BEV Padding, the rotation error can improve by 1.41° (IDs 0 and 1).

\section{Conclusions}
We introduced VVLoc, a unified approach for both topological and metric localization in multi-camera systems. 
VVLoc addresses key limitations of prior methods, including reliance on pose or semantic priors and the absence of confidence measurements. 
In addition to public datasets, we collected a challenging real-world dataset to evaluate global localization, loop closure detection, pose estimation, and multi-modal mapping. 
Experimental results demonstrate VVLoc's state-of-the-art performance across various tasks, offering a scalable and resource-efficient solution for real-world deployment. 
As a result, VVLoc has been successfully deployed in mass production, proving its practicality for autonomous driving and intelligent transportation systems.

\bibliographystyle{plain}
\bibliography{ref}

@article{zhang2022bev,
  title={Bev-locator: An end-to-end visual semantic localization network using multi-view images},
  author={Zhang, Zhihuang and Xu, Meng and Zhou, Wenqiang and Peng, Tao and Li, Liang and Poslad, Stefan},
  journal={Science China Information Sciences},
  year={2025},
}

@inproceedings{camiletto2023u,
  title={U-bev: Height-aware bird’s-eye-view segmentation and neural map-based relocalization},
  author={Camiletto, Andrea Boscolo and Bochicchio, Alfredo and Liniger, Alexander and Dai, Dengxin and Gawel, Abel},
  booktitle={IROS},
  year={2024},
}

@inproceedings{he2023egovm,
  title={Egovm: Achieving precise ego-localization using lightweight vectorized maps},
  author={He, Yuzhe and Liang, Shuang and Rui, Xiaofei and Cai, Chengying and Wan, Guowei},
  booktitle={IROS},
  year={2024},
}

@inproceedings{shore2023bev,
  title={BEV-CV: Birds-Eye-View Transform for Cross-View Geo-Localisation},
  author={Shore, Tavis and Hadfield, Simon and Mendez, Oscar},
  booktitle={IROS},
  year={2024},
}

@inproceedings{zhao2022jperceiver,
  title={Jperceiver: Joint perception network for depth, pose and layout estimation in driving scenes},
  author={Zhao, Haimei and Zhang, Jing and Zhang, Sen and Tao, Dacheng},
  booktitle={ECCV},
  year={2022}
}

@article{fervers2023c,
  title={C-BEV: Contrastive Bird's Eye View Training for Cross-View Image Retrieval and 3-DoF Pose Estimation},
  author={Fervers, Florian and Bullinger, Sebastian and Bodensteiner, Christoph and Arens, Michael and Stiefelhagen, Rainer},
  journal={arXiv},
  year={2023}
}

@article{xu2023leveraging,
  title={Leveraging BEV Representation for 360-degree Visual Place Recognition},
  author={Xu, Xuecheng and Jiao, Yanmei and Lu, Sha and Ding, Xiaqing and Xiong, Rong and Wang, Yue},
  journal={arXiv},
  year={2023}
}

@article{ge2024bev2pr,
  title={BEV2PR: BEV-Enhanced Visual Place Recognition with Structural Cues},
  author={Ge, Fudong and Zhang, Yiwei and Shen, Shuhan and Wang, Yue and Hu, Weiming and Gao, Jin},
  journal={IROS},
  year={2024}
}

@article{li2023occ,
  title={OCC-VO: Dense Mapping via 3D Occupancy-Based Visual Odometry for Autonomous Driving},
  author={Li, Heng and Duan, Yifan and Zhang, Xinran and Liu, Haiyi and Ji, Jianmin and Zhang, Yanyong},
  journal={ICRA},
  year={2024}
}

@article{wu2024maplocnet,
  title={MapLocNet: Coarse-to-Fine Feature Registration for Visual Re-Localization in Navigation Maps},
  author={Wu, Hang and Zhang, Zhenghao and Lin, Siyuan and Mu, Xiangru and Zhao, Qiang and Yang, Ming and Qin, Tong},
  journal={IROS},
  year={2024}
}

@article{carlevaris2016university,
  title={University of Michigan North Campus long-term vision and lidar dataset},
  author={Carlevaris-Bianco, Nicholas and Ushani, Arash K and Eustice, Ryan M},
  journal={IJRR},
  year={2016}
}

@inproceedings{huang2021predator,
  title={Predator: Registration of 3d point clouds with low overlap},
  author={Huang, Shengyu and Gojcic, Zan and Usvyatsov, Mikhail and Wieser, Andreas and Schindler, Konrad},
  booktitle={CVPR},
  year={2021}
}

@inproceedings{chen2022sc2,
  title={Sc2-pcr: A second order spatial compatibility for efficient and robust point cloud registration},
  author={Chen, Zhi and Sun, Kun and Yang, Fan and Tao, Wenbing},
  booktitle={CVPR},
  year={2022}
}

@article{arun1987least,
  title={Least-squares fitting of two 3-D point sets},
  author={Arun, K Somani and Huang, Thomas S and Blostein, Steven D},
  journal={TPAMI},
  year={1987},
  publisher={IEEE}
}

@inproceedings{li2022bevformer,
  title={Bevformer: Learning bird’s-eye-view representation from multi-camera images via spatiotemporal transformers},
  author={Li, Zhiqi and Wang, Wenhai and Li, Hongyang and Xie, Enze and Sima, Chonghao and Lu, Tong and Qiao, Yu and Dai, Jifeng},
  booktitle={ECCV},
  year={2022},
}

@inproceedings{arandjelovic2016netvlad,
  title={NetVLAD: CNN architecture for weakly supervised place recognition},
  author={Arandjelovic, Relja and Gronat, Petr and Torii, Akihiko and Pajdla, Tomas and Sivic, Josef},
  booktitle={CVPR},
  year={2016}
}

@inproceedings{jin2017learned,
  title={Learned contextual feature reweighting for image geo-localization},
  author={Jin Kim, Hyo and Dunn, Enrique and Frahm, Jan-Michael},
  booktitle={CVPR},
  year={2017}
}

@article{radenovic2018fine,
  title={Fine-tuning CNN image retrieval with no human annotation},
  author={Radenovi{\'c}, Filip and Tolias, Giorgos and Chum, Ond{\v{r}}ej},
  journal={TPAMI},
  year={2018},
}

@inproceedings{detone2018superpoint,
  title={Superpoint: Self-supervised interest point detection and description},
  author={DeTone, Daniel and Malisiewicz, Tomasz and Rabinovich, Andrew},
  booktitle={CVPRW},
  year={2018}
}

@inproceedings{edstedt2024roma,
  title={RoMa: Robust dense feature matching},
  author={Edstedt, Johan and Sun, Qiyu and B{\"o}kman, Georg and Wadenb{\"a}ck, M{\aa}rten and Felsberg, Michael},
  booktitle={CVPR},
  year={2024}
}

@inproceedings{hausler2021patch,
  title={Patch-netvlad: Multi-scale fusion of locally-global descriptors for place recognition},
  author={Hausler, Stephen and Garg, Sourav and Xu, Ming and Milford, Michael and Fischer, Tobias},
  booktitle={CVPR},
  year={2021}
}

@article{tzachor2024effovpr,
  title={EffoVPR: Effective Foundation Model Utilization for Visual Place Recognition},
  author={Tzachor, Issar and Lerner, Boaz and Levy, Matan and Green, Michael and Shalev, Tal Berkovitz and Habib, Gavriel and Samuel, Dvir and Zailer, Noam Korngut and Shimshi, Or and Darshan, Nir and others},
  journal={arXiv},
  year={2024}
}

@inproceedings{lu2024cricavpr,
  title={CricaVPR: Cross-image Correlation-aware Representation Learning for Visual Place Recognition},
  author={Lu, Feng and Lan, Xiangyuan and Zhang, Lijun and Jiang, Dongmei and Wang, Yaowei and Yuan, Chun},
  booktitle={CVPR},
  year={2024}
}

@inproceedings{edstedt2023dkm,
  title={DKM: Dense kernelized feature matching for geometry estimation},
  author={Edstedt, Johan and Athanasiadis, Ioannis and Wadenb{\"a}ck, M{\aa}rten and Felsberg, Michael},
  booktitle={CVPR},
  year={2023}
}

@inproceedings{barroso2024matching,
  title={Matching 2D Images in 3D: Metric Relative Pose from Metric Correspondences},
  author={Barroso-Laguna, Axel and Munukutla, Sowmya and Prisacariu, Victor Adrian and Brachmann, Eric},
  booktitle={CVPR},
  year={2024}
}

@inproceedings{barroso2019key,
  title={Key. net: Keypoint detection by handcrafted and learned cnn filters},
  author={Barroso-Laguna, Axel and Riba, Edgar and Ponsa, Daniel and Mikolajczyk, Krystian},
  booktitle={ICCV},
  year={2019}
}

@inproceedings{gleize2023silk,
  title={SiLK: Simple Learned Keypoints},
  author={Gleize, Pierre and Wang, Weiyao and Feiszli, Matt},
  booktitle={ICCV},
  year={2023}
}

@article{revaud2019r2d2,
  title={R2d2: Reliable and repeatable detector and descriptor},
  author={Revaud, Jerome and De Souza, Cesar and Humenberger, Martin and Weinzaepfel, Philippe},
  journal={NeurIPS},
  year={2019}
}

@inproceedings{jiang2021cotr,
  title={Cotr: Correspondence transformer for matching across images},
  author={Jiang, Wei and Trulls, Eduard and Hosang, Jan and Tagliasacchi, Andrea and Yi, Kwang Moo},
  booktitle={ICCV},
  year={2021}
}

@inproceedings{sun2021loftr,
  title={LoFTR: Detector-free local feature matching with transformers},
  author={Sun, Jiaming and Shen, Zehong and Wang, Yuang and Bao, Hujun and Zhou, Xiaowei},
  booktitle={CVPR},
  year={2021}
}

@inproceedings{cai2021extreme,
  title={Extreme rotation estimation using dense correlation volumes},
  author={Cai, Ruojin and Hariharan, Bharath and Snavely, Noah and Averbuch-Elor, Hadar},
  booktitle={CVPR},
  year={2021}
}

@inproceedings{winkelbauer2021learning,
  title={Learning to localize in new environments from synthetic training data},
  author={Winkelbauer, Dominik and Denninger, Maximilian and Triebel, Rudolph},
  booktitle={ICRA},
  year={2021},
}

@inproceedings{tolias2020learning,
  title={Learning and aggregating deep local descriptors for instance-level recognition},
  author={Tolias, Giorgos and Jenicek, Tomas and Chum, Ond{\v{r}}ej},
  booktitle={ECCV},
  year={2020},
}

@inproceedings{yang2021dolg,
  title={Dolg: Single-stage image retrieval with deep orthogonal fusion of local and global features},
  author={Yang, Min and He, Dongliang and Fan, Miao and Shi, Baorong and Xue, Xuetong and Li, Fu and Ding, Errui and Huang, Jizhou},
  booktitle={ICCV},
  year={2021}
}

@article{qin2023geotransformer,
  title={Geotransformer: Fast and robust point cloud registration with geometric transformer},
  author={Qin, Zheng and Yu, Hao and Wang, Changjian and Guo, Yulan and Peng, Yuxing and Ilic, Slobodan and Hu, Dewen and Xu, Kai},
  journal={TPAMI},
  year={2023},
}

@inproceedings{sun2020circle,
  title={Circle loss: A unified perspective of pair similarity optimization},
  author={Sun, Yifan and Cheng, Changmao and Zhang, Yuhan and Zhang, Chi and Zheng, Liang and Wang, Zhongdao and Wei, Yichen},
  booktitle={CVPR},
  year={2020}
}

@article{chen20203d,
  title={3D Lidar Mapping Relative Accuracy Automatic Evaluation Algorithm},
  author={Chen, Guibin and Deng, Jiong and Huang, Dongze and Zhang, Shuo},
  journal={arXiv},
  year={2020}
}

@article{RobotCarDatasetIJRR, 
  Author = {Will Maddern and Geoff Pascoe and Chris Linegar and Paul Newman}, 
  Title = {{1 Year, 1000km: The Oxford RobotCar Dataset}}, 
  Journal = {IJRR}, 
  Year = {2017}
}

\end{document}